# English Character Recognition using Artificial Neural Network

TIRTHARAJ DASH
Department of Information Technology
National Institute of Science and Technology
Berhampur-761008, India
Email: tirtharajnist446@gmail.com

TANISTHA NAYAK
Department of Information Technology
National Institute of Science and Technology
Berhampur-761008, India
Email: tanisthanist213@gmail.com

*Abstract*— **This work focuses on development of a Offline Hand Written English Character Recognition algorithm based on Artificial Neural Network (ANN). The ANN implemented in this work has single output neuron which shows whether the tested character belongs to a particular cluster or not. The implementation is carried out completely in 'C' language. Ten sets of English alphabets (small-26, capital-26) were used to train the ANN and 5 sets of English alphabets were used to test the network. The characters were collected from different persons over duration of about 25 days. The algorithm was tested with 5 capital letters and 5 small letter sets. However, the result showed that the algorithm recognized English alphabet patterns with maximum accuracy of 92.59% and False Rejection Rate (FRR) of 0%.**

## I. INTRODUCTION

In this machine learning world, English Hand Written Character Recognition has been a challenging and interesting research area in the field of Artificial Intelligence and Soft computing [1,2]. It contributes majorly to the Human and Computer interaction and improves the interface between the two [3]. Other human cognition methods viz. face, speech, thumb print recognitions are also being great area of research [4,5,6].

Generally, character recognition can be broadly characterized into two types (i) Offline and (ii) Online. In offline method, the pattern is captured as an image and taken for testing purpose. But in case of online approach, each point of the pattern is a function of pressure, time, slant, strokes and other physical parameters. Both the methods are best based on their application in the field of machine learning. Yielding best accuracy with minimal cost of time is a crucial precondition for pattern recognition system. Therefore, hand written character recognition is continuously being a broad area of research.

In this work, an approach for offline English character recognition has been proposed using Artificial Neural Network (ANN). ANN basically resembles with the characteristics of a Biological Neural Net (BNN). Knowledge bases of the ANN are the inter-neural weights. Based on these two facts, we developed one algorithm for hand written character recognition.

This paper is organized as follows. Section 1 presented a general introduction to the character recognition systems and methods. Section 2 gives a brief survey of some methods proposed for character recognition. Section 3 describes the proposed methodology of this work. Section 4 is a result and discussion section which gives a detailed analysis of the work. The paper is concluded in section 5 with a note to future works.

## II. LITERATURE SURVEY

Available literatures convey that various approaches have been made in order to accomplish the task of character recognition. And in about all the soft computing approaches Neural Network (NN) has been a backend of character classification. This is due to its faster computation. The methods used in front end could be (a) statistical approaches (b) kernel methods (c) support methods or (d) hybrid of fuzzy logic controllers.

### A. Use of MLP

Multilayer Perceptron (MLP) was used for 'Bangla' alphabet recognition by [8]. The accuracy achieved in this work was 86.46% and 75.05% on the samples of training and testing respectively.

[9] proposed NN based English character recognition system. In this work, MLP with one hidden layer was used. About 500 testing were carried out to test the performance of the design. The best case accuracy obtained in this work was 94%.

### B. Statistical and NN

[10] used horizontal and vertical strokes and end points as feature for handwritten numerals. This method reported an accuracy rate of 90.5% in best case. However, this method used a thinning method resulting in loss of features.

[11] worked on English alphabet recognition using NN. In this work, binary pixels of the alphabets were used train the NN. The accuracy achieved was found to be 82.5%. These technique can also be applied to any pattern recognition with different neural network algorithms [12].

A detailed analysis of some methods is given in Table-1 below which shows the References, approach and its corresponding accuracy.

**TABLE-1** Some Approaches with their Performance

| Approach | Reference | Rate of Recognition (%) |
|---|---|---|
| Multilayer Perceptron (MLP) | [8] | 75.05% |
| MLP with NN | [9] | 94% |
| Stroke Method | [10] | 90.5% |
| Neural Network with Statistical Approach | [11] | 82.5% |

### III. METHODOLOGY

The steps followed in this work of English character recognition are given in the algorithm below.

**Algorithm:**
*Step 0:* **START**
*Step 1:* Collect the Training English Characters (10 sets × 52)
*Step 2:* Collect the Test Patterns (5 sets × 52)
*Step 3:* Extract the features from patterns collected in Step-1 and 2.
*Step 4:* Develop of the ANN algorithm for Training and Testing using 'C'
*Step 5:* Train the developed net with training patterns and store the weight (knowledge base).
*Step 6:* **Repeat** Step-5 if you want to reduce the Error.
*Step 7:* Perform the testing for the testing patterns and store the weights.
*Step 8:* Check the updated knowledge base with all the previous knowledge bases.
*Step 9:* **STOP** with an accuracy percentage

The updation of knowledge is given by Equation-1 below.

$$K(new) = K(old) + Input \times Target \quad (1)$$

The Feature Extraction of the pattern can be demonstrated as in the Figure-1 below.

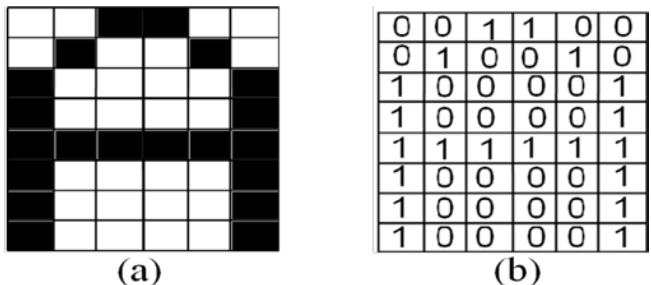

**Figure-1 (a)** English Character Pattern **(b)** Extracted Pixels

The method proposed in Algorithm above is demonstrated in Figure-2 below.

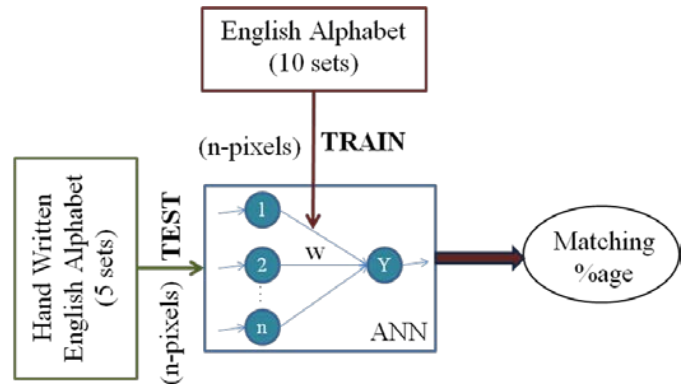

**Figure-2** Proposed Methodology

### IV. RESULT

It should be mentioned that the proposed algorithm is currently under the rigorous testing by the authors. However, the results discussed below out perform some of the mentioned techniques in the Literature survey section.

We have tested our developed algorithm with 5 capital letters and 5 small English character alphabets. The result is given in Table-2.

TABLE-2 Result shows the performance of the developed algorithm

| Input Pattern | Training Count | Testing Pattern | Testing Count | Unrecognized | Success Rate (%) |
|---|---|---|---|---|---|
| A | 10 | A | 5 | 0 | 82.08 |
| B | 10 | B | 5 | 2 | 68.78 |
| C | 10 | C | 5 | 0 | 91.34 |
| D | 10 | D | 5 | 3 | 54.89 |
| E | 10 | E | 5 | 2 | 65.00 |
| a | 10 | a | 5 | 1 | 83.45 |
| b | 10 | b | 5 | 2 | 77.09 |
| c | 10 | c | 5 | 0 | 92.59 |
| d | 10 | d | 5 | 4 | 34.81 |
| e | 10 | e | 5 | 3 | 53.20 |

A plot has been given in Figure-3 below to view the performance of the developed algorithm with respect to its rate of recognition of patterns.

Two main parameters decide the performance of the algorithm.
1. Success Rate
2. False Rejection Rate (FRR)

The success rate defines the rate of recognition whereas the FRR defines the ratio between the unrecognized patterns and the total number of testing patterns.

## V. CONCLUSION

This work proposed an algorithm for hand written English alphabet pattern recognition. The algorithm is based on principle of Artificial Neural Network (ANN). However, the algorithm was tested with 5 capital letters set and 5 small letters set. The result and discussion section showed that the developed algorithm works well with maximum recognition rate of 92.59% and minimum False Rejection Rate (FRR) of 0%.

As future work, it will be interesting to test the developed algorithm with more number of English alphabet samples. The authors are currently working on this part.

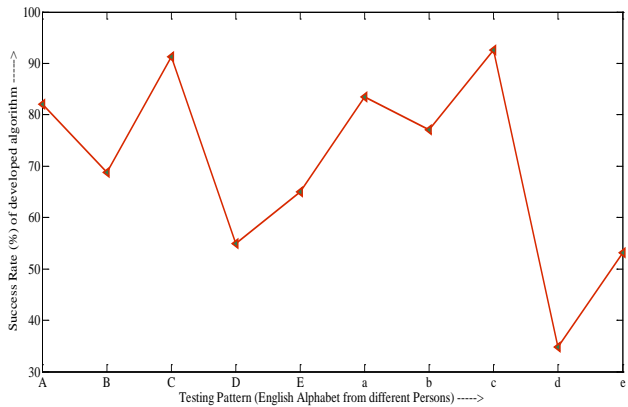

**Figure-3:** Testing pattern vs. success rate

False Rejection Rate can be calculated as in Equation-2 below.

$$\text{FRR} = \frac{\text{Number of Patterns Unrecognized}}{\text{Total Number of Patterns Tested}} \times 100\% \quad (2)$$

A plot showing testing pattern vs. FRR is given in Figure-4 below.

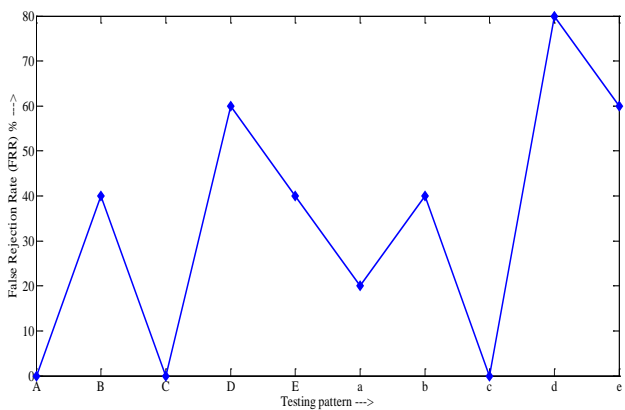

**Figure-4:** Testing pattern vs. FRR

The plot (Figure-5) shows the relative success rate versus the false rejection rate of the tested patterns.

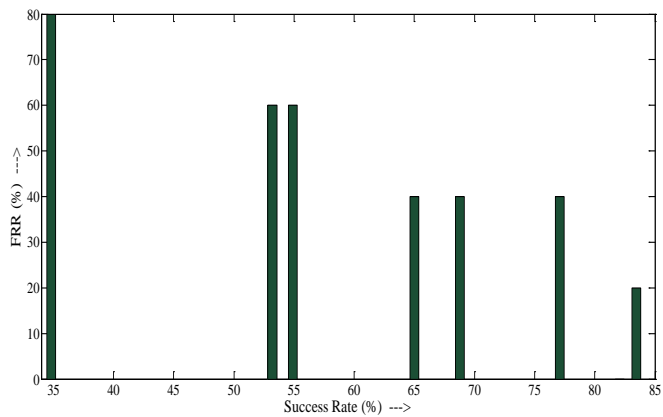

**Figure-5:** Success rate vs. FRR